\documentclass[journal,10pt]{IEEEtran} 

\usepackage[utf8]{inputenc}

\ifCLASSINFOpdf
\else
\fi

\usepackage{amsmath,graphicx}
\usepackage{amsfonts}
\usepackage{amssymb}
\usepackage{float}
\usepackage{hyperref}
\newlength\figureheight
\newlength\figurewidth
\usepackage{color}

\def\x{{\mathbf x}}

\def\baseGP{f}
\def\R{\mathbb{R}}
\def\a{\mathbf{a}}

\def\m{\mathbf{m}}
\def\W{\mathbf{W}}
\def\w{\mathbf{w}}

\def\y{\mathbf{y}}

\def\cD{\mathcal{D}}

\def\cN{\mathcal{N}}
\def\meas{m_{i,j}}
\def\loc{x_{i,j}}
\def\measNse{\epsilon_{i,j}}
\def\obsNse{\eta_i}
\def\wgt{w_{i,j}}
\def\obs{y_i}

\def\weight{w}
\newcommand{\weightElem}[1]{{\weight}_{#1}}

\def\x{\mathbf{x}}

\def\observation{y}

\def\expectSymb{\mathbb{E}}
\def\location{\mathbf{x}}
\newcommand{\expectation}[1]{\expectSymb \left(#1\right)}

\begin{document}
%
\title{Recovering Latent Signals from a Mixture of Measurements using a Gaussian Process Prior}

\author{Felipe Tobar, Gonzalo Rios, Tomás Valdivia and Pablo Guerrero
\thanks{
F.T and G.R. and T.V. are with the Center for Mathematical Modeling (CMM), Universidad de Chile, and P.G is with the Computer Science Department and the Advanced Mining Technology Center, Universidad de Chile. F.T. thanks CONICYT PAI-82140061 and Basal-CMM, P.G. to FONDECYT Postdoctoral 3130729, and G.R. to CONICYT-PCHA Doctorado Nacional 2016-21161789.}
}

\markboth{Author copy submitted to IEEE Signal Processing Letters}%
{Tobar \MakeLowercase{\textit{et al.}}: GPMM}

\maketitle

\begin{abstract}
In sensing applications, sensors cannot always measure the latent quantity of interest at the required resolution, sometimes they can only acquire a blurred version of it due the sensor's transfer function. To recover latent signals when only noisy mixed measurements of the signal are available, we propose the Gaussian process mixture of measurements (GPMM), which models the latent signal as a Gaussian process (GP) and allows us to perform Bayesian inference on  such signal conditional to a set of noisy mixture of measurements. We describe how to train GPMM, that is, to find the hyperparameters of the GP and the mixing weights, and how to perform inference on the latent signal under GPMM; additionally, we identify the solution to the underdetermined linear system resulting from a sensing application as a particular case of GPMM. The proposed model is validated in the recovery of three signals: a smooth synthetic signal, a real-world heart-rate time series and a step function, where GPMM outperformed the standard GP in terms of estimation error, uncertainty representation and recovery of the spectral content of the latent signal.
\end{abstract}
\begin{IEEEkeywords}
  Gaussian process, convolution process, sensing applications, Bayesian inference.
\end{IEEEkeywords}

\IEEEpeerreviewmaketitle

\section{Introduction}
\label{sec:intro}
\begin{figure}[t]
   \centering
  \includegraphics[width=0.5\textwidth]{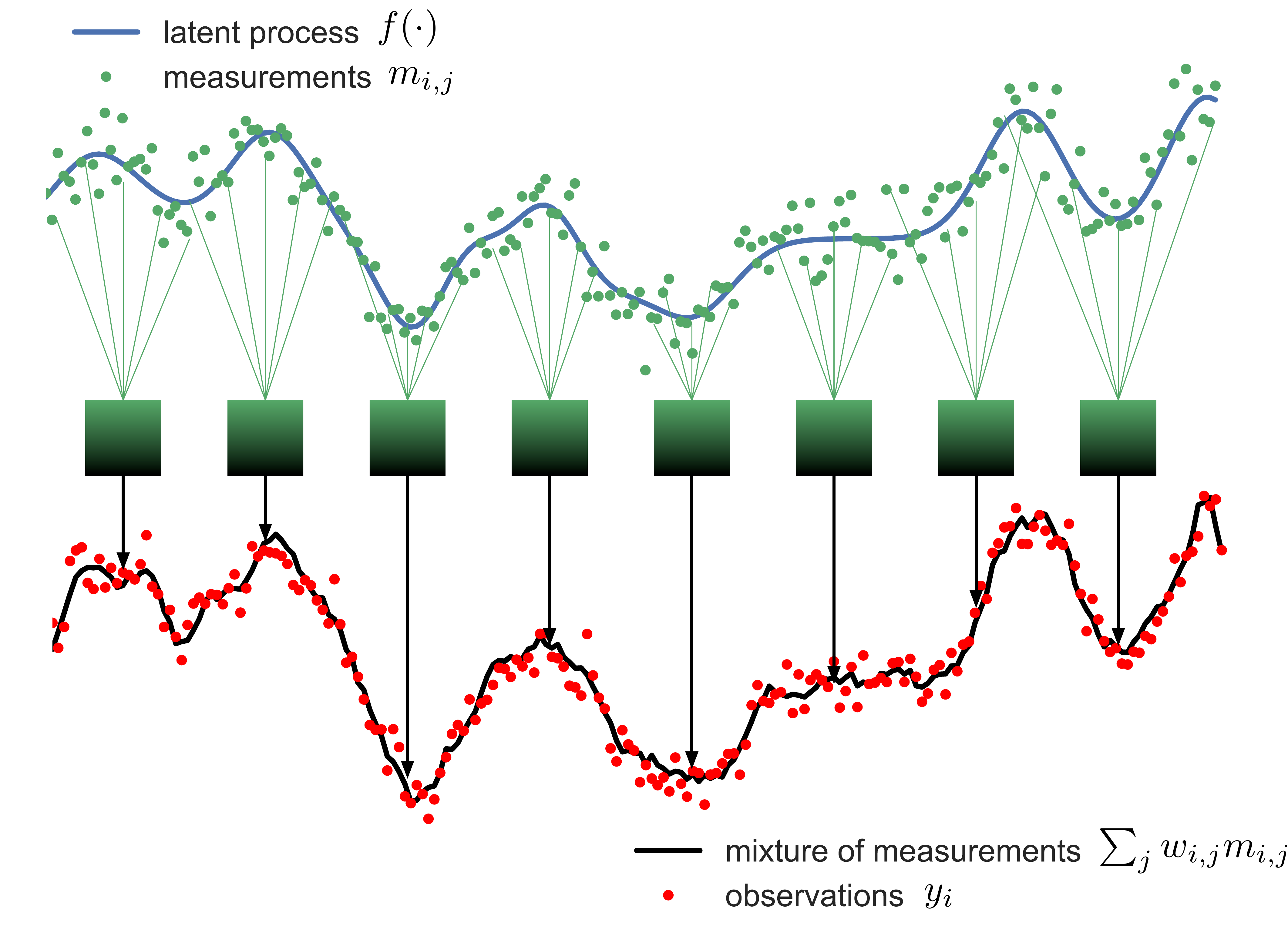}
  \caption{Sensing application:  A latent signal (blue) is measured
up to sensor noise (green), these measurements are linearly mixed to yield the mixture-of-observations process (black) which is in turn corrupted by noise (red).}
    \label{fig:diagram}
\end{figure}

Observations within sensing applications result from the convolution between the latent signal and the sensors's transfer function, therefore, a desired property of the sensor is to have a transfer function that is close to a Dirac delta function so that the latent signal can be recovered from the observations. We will model this convolution in a discrete manner to give rise to the representation of a sensing application described in fig. \ref{fig:diagram}, where we model the observations as a (noisy) mixture of (again noisy) measurements and aim to recover the latent signal from the observations. Mixing of the latent signal's values stems from the inability of the sensor to measure the latent signal at the required resolution, this is due to low quality of the sensors that \textit{colour} the observations which have to then be \textit{whiten} in order to recover the latent process.  Observations composed by mixtures of measurements are commonplace in sensing applications in different areas: in robot localization using radars or sonars \cite{adams_laser,Moravec85}, in astronomical applications \cite{starck2002deconvolution}, and in super-resolution image recovery \cite{baboulaz2009}, to name but a few.

 A workaround to the problem of recovering a latent process from observations composed by mixtures of measurements is to define a set of sensing locations (i.e., a grid) and model the observations as a system of linear equations---recall that the measurements are combined in a linear fashion.However, this approach is rather restrictive since it constrains the measurements to be collected at fixed locations, which is rarely the case in real-world applications, and it leads to heavily underdetermined linear systems. In our view, the key to this problem is to assume spatial correlations in the latent signals so that observations of overlapping regions reveal structure in the signal; we do so in a probabilistic manner by placing a Gaussian process (GP) prior \cite{rasmussen06} on the signal, which is then updated into the posterior density of the latent signal conditional to such observations. However, GPs in their standard form are not suited to deal with observations comprising a mixture of measurements.

In this letter we propose a GP-based mixture-of-measurements model, illustrate how to train it and perform Bayesian inference on the latent signal, establish a connection between the proposed model and the linear system approach to sensing applications, and validate our model on real-world and synthetic data

\subsection{Background: Gaussian processes and related work}

Gaussian processes (GPs) \cite{rasmussen06} are a nonparametric prior distribution on functions $f:X\mapsto\R$, for any set $X$ where a covariance function can be defined (e.g., a metric space), such that for any finite collection of inputs $\x=[x_1,x_2,\ldots,x_n]^\top \in X^n$, the corresponding outputs $f(x_1),f(x_2),\ldots,f(x_n)\in \R$ are jointly Gaussian, that is,
\begin{equation}
[f(x_1),f(x_2),\ldots,f(x_n)]^\top\sim \cN(\mu(\x),K(\x,\x))
\end{equation}
where the mean and covariance {functions}, $\mu(\cdot)$ and $K(\cdot,\cdot)$ respectively, are parametric forms that determine the spatial properties of samples drawn from the GP prior. Training the GP to observed data involves finding the parameters of the mean and covariance functions, then, the \textit{posterior} distribution of the entire function $f$ conditional to a set of observations is Gaussian.

The GP framework is well suited to the sensing setting in fig. \ref{fig:diagram}, since modelling latent signal as a GP results in the posterior distribution of the latent process (given the mixture of measurements) being Gaussian as well. Previous GP-based models for convolution processes \cite{nips15,npr_15b,boyle,Higdon2002} model signals a convolution between a continuous-time filter and a white-noise process, which is unsuitable to represent the latent process in the sensing application where the spatial correlation of the process is fundamental. Conversely,  \cite{NIPS2008_3553} allows GPs as latent functions but to address the multi-output case, where the aim is to perform inference on the outputs rather than the latent processes. Furthermore, these methods consider continuous-time convolution filters, which is computationally demanding and requires, e.g., variational approximations \cite{titsias09,chatzis15}. Consequently, closed-form and computationally-efficient Bayesian reconstruction of the latent process is still an open problem  in sensing applications.

\section{The Gaussian Process Mixture of Measurements (GPMM)}

Consider a sensing application where each observation $y_i$ is a noisy mixture of, again noisy and hidden, measurements $m_{i,j}$ of a latent process $f(\cdot)$ measured at locations $x_{i,j}$, that is,
\begin{align}
m_{i,j} &= f(x_{i,j}) + \epsilon_{i,j} \label{eq:mix1a}\\
y_i&= \sum_{j=1}^{M}w_{i,j}m_{i,j}+\eta_{i} \label{eq:mix1b}
\end{align}
where for the $i^{\text{th}}$ observation $\obs$, we use the following notation:
\begin{itemize}
\item $f(\loc)$ is the value of the latent process at location $\loc$,

\item $\meas$ is the measurement of $f(x_{i,j})$ acquired by the sensor,
\item $\wgt$ is the weight of the measurement $\meas$,
\item $M$ is the number of measurements in the observation $\obs$,
\item $\measNse$ is the measurement  Gaussian noise, and
\item $\obsNse$ is the observation Gaussian noise.
\end{itemize}
We use the terms \textit{measurement} and \textit{observation} defined here consistently in the rest of the paper---see fig. \ref{fig:diagram}. Note that in general, different observations, i.e. $y_{i}, y_{i'}\ i\neq i'$, correspond to different regions and both the locations $x_{i,j},x_{i',j'},  \forall i,j,j'$ and the weights $w_{i,j},w_{i',j'}, \forall i,j,j'$ might be different due to the sensing procedure.

For $N$ observations, we express eq. \eqref{eq:mix1b} as a system of linear equations in block matrix notation:

\begin{equation}
\underbrace{\left[\begin{smallmatrix} y_{1} \\ \vdots \\ y_{N} \end{smallmatrix}\right]}_{\y}=
 {\underbrace{\left[\begin{smallmatrix} \w_{1} &  & \\
& \ddots  &  \\
           & & \w_{N}\end{smallmatrix}\right]}_{\W}}^T
\underbrace{\left[\begin{smallmatrix} \m_{1} \\
        \vdots \\
         \m_{M}\end{smallmatrix}\right]}_{\m}
+
\underbrace{\left[\begin{smallmatrix} \eta_{1} \\ \vdots \\ \eta_{N} \end{smallmatrix}\right]}_{\pmb{\eta}}\label{eq:matrix_system}
\end{equation}
where $\w_{i} = [w_{i,1},\ldots,w_{i,M}]^T$, $\m_{i} = [m_{i,1},\ldots,m_{i,M}]^T$ and the matrix $\W$ is an $M$-diagonal matrix. Column vectors are denoted in bold lowercase font and matrices in bold uppercase font.

\subsection{Solving the linear system}

Our approach will consider the weights $w_{i,j}$ to be unknown and learn them from the observations, however, let us assume they are known in this section in order to address the problem as a linear system; for ease of notation, we assume there are no common locations across the measurements (i.e., $ i\neq i', j\neq j' \Rightarrow x_{i,j}\neq x_{i',j'}$). With these assumptions, eq. \eqref{eq:matrix_system} is an underdetermined linear system: there are $L =MN$ unknowns and $N$ equations, where  $N\ll L$, in fact, the general solution to such a system has $L-N$ free parameters or degrees of freedom (neglecting the inconsistent case).

This underdetermined system has infinite solutions, with the minimum-norm solution given by \mbox{$\hat{\m} = \W^+\y$}, where $\W^+$ is the Moore-Penrose pseudoinverse of $\W$ \cite{Moore1920}. Using this solution to recover the latent signal has a number of drawbacks: (i) it requires the weights $w_{i,j}$ to be known, (ii) it only recovers the process at the measured locations without providing any insight about regions not measured, and (iii) it does not provide a measure of uncertainty for the estimates, e.g., in the form of error bars.

\subsection{A novel generative model for the mixture of measurements}

As the system in eq. \eqref{eq:matrix_system} has infinite solutions, a constraint (or regularisation criterion) has to be imposed to reduce the number of solutions, or critically, to find a single solution. The Moore-Penrose pseudoinverse imposes the minimum-norm criterion, we instead assume a probabilistic condition (a prior) on the spatial structure of the solution. Specifically, we (i) place a GP prior on the latent signal to then (ii) find the posterior distribution of the latent signal conditional to the mixture of measurements, even at locations that were not measured. A key property of this approach is that  these two steps are performed analytically, since the latent signal and the mixture of measurements are jointly Gaussian. We next present a formal description of the proposed generative model.

We model the latent process $\baseGP(\cdot)$ in \eqref{eq:mix1a} as a GP over the set of locations $X$ given by
\begin{equation}
  \label{eq:latent}
  f\sim\ \mathcal{GP}(\mu_f,K_f)
\end{equation}
where $\mu_f:X\mapsto\R$ and $K_f:X\times X\mapsto\R$ are the GP mean and covariance functions respectively. As the linear combination of jointly-Gaussian random variables (RVs) is Gaussian, the observations in eq.~\eqref{eq:mix1b} are Gaussian RVs indexed by
$\x_i=[x_{i,1},\ldots,x_{i,M}]^\top\in X^{M}$
with mean and covariance respectively given by
\begin{align}
  \mu_y(\x_i) &=
  \label{eq:mean_mix}
  \expectation{\observation_i} = \sum_{j=1}^{M} \weightElem{i,j} \mu_f(x_{i,j})\\
  \label{eq:K_mix}
  K_{y}(\x_i, \x_{i'}) &= \expectation{(\observation_i-\mu_y(\x_i))(\observation_{i'}-\mu_y (\x_{i'})) } \\
  & \hspace{-4.5em}= \sum_{j,j'=1}^{M} \weightElem{i,j}\weightElem{i',j'}\left(K_f(x_{i,j},x_{i',j'}) +\sigma^2_\epsilon\delta(x_{i,j}-x_{i',j'})\right) + \delta_{i-i'}\sigma^2_\eta \nonumber
\end{align}
where $\sigma^2_\eta,\sigma^2_\epsilon$ are the variances of the measurement and observation noises respectively. Additionally, note that if measurements are always taken at different locations (which is the case if the set of locations is continuous) we have $\delta(x_{i,j}-x_{i',j'})=0\ \forall i , i', j ,j'$ and
\begin{align}
K_{y}(\x_i, \x_{i'})  &= \sum_{j=1}^{M}\sum_{j'=1}^{M} \weightElem{i,j}\weightElem{i',j'} K_f(x_{i,j},x_{i',j'}) + \delta_{i-i'}\sigma^2_\eta\\
&=\w_i^\top{K_f}(\x_i,\x_{i'}) \w_{i'} + \delta_{i-i'}\sigma^2_\eta\label{eq:Ky}
\end{align}
where ${K_f}(\x_i,\x_{i'})\in\R^{M\times M}$ is the Gram matrix evaluated on $\x_i $ and $ \x_{i'}$, $\x_i=[x_{i,1},\ldots,x_{i,M}]^\top$ and $\w_i=[w_{i,1},\ldots,w_{i,M}]^\top$.

\begin{figure}[t]
   \centering
  \includegraphics{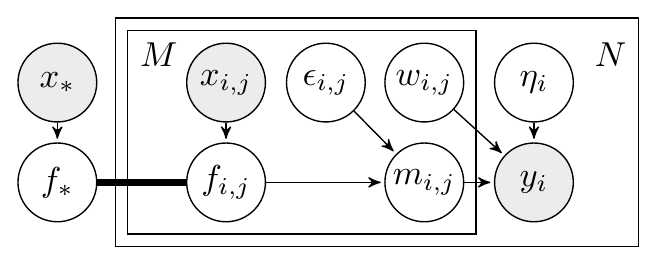}
  \caption{Graphical model of GPMM. Hidden variables are left blank whereas observed ones are shaded. The inner plate represents the $(i,j)^\text{th}$ measurement, the outer one the $i^\text{th}$ observation, and $x_*$ and $f_*$ are a test location and its value respectively. The thick bar indicates that  the latent GP $f$ is completely interconnected}
    \label{fig:LCOGPGraphicalModel}
\end{figure}

The covariance of the observations is a mixture of evaluations of the covariance function of the process $f(\cdot)$ at the locations measured. This implies that a single entry in common between measurement locations $\x_i$ and $\x_j$ is enough for the observations $y_i(\x_i)$ and $y_j(\x_j)$ to be correlated. This mixture-of-kernel structure  resembles additive GPs  \cite{Duvenaud11} and multikernel learning \cite{gonen11,tobar14, Yukawa12}, however, these methods combine different kernels (evaluated on a common input) for expressive kernel design, whereas the proposed model combines evaluations of a single kernel on different locations to find the spatial structure of the latent signal. We refer to the presented model as Gaussian processes mixture of measurements (GPMM) and give a graphical model illustration in fig. \ref{fig:LCOGPGraphicalModel}.

\section{Inference for GPMM and Relationship to the Moore-Penrose Pseudoinverse}

Fitting the model to the observations $\cD=\{(\location_i, \observation_i), i=1,\ldots,N\}$ involves finding appropriate hyperparameters for GPMM, $\theta_{\text{GPMM}}$, that is, the hyperparameters of the latent signal in   eq. \eqref{eq:latent} and the weights $w_i$ which are now hyperparameters. This can be achieved by minimising the negative log-likelihood $\log p(\y)$:
\begin{align*}
    \theta_{\text{GPMM}}=\arg\min\frac{1}{2}\y^\top K_y^{-1}\y + \frac{1}{2}\log{\vert K_y \vert} + \frac{N}{2}\log{2\pi}
\end{align*}
where $K_y$ is given in \eqref{eq:K_mix}, $\y=[y_1,\ldots,y_N]^\top$, and the optimisation can be performed by, e.g., gradient descent. Notice that the cost of computing $K_y$ in eq. \eqref{eq:Ky} and inverting it are respectively $\mathcal{O}(N^2M^2)$ and $\mathcal{O}(N^3)$, meaning that if $M$ is dominated by $\mathcal{O}(N^{1/2})$ the cost of training GPMM is $\mathcal{O}(N^3)$ as standard GPs. With the optimal hyperparameters, we are now ready to calculate the posterior of $f(\cdot)$.

\subsection{The posterior of the latent process}
\label{sec_posterior}
The posterior density of $f$ given the observations $\cD$, $p(f \vert \cD)$, is Gaussian and determined by (i) the prior mean---assumed to be zero in this case, (ii) the  autocovariance of the mixture-of-measurement process $y$---given in eq.~\eqref{eq:K_mix}, and (iii) the covariance between the latent process $f(x)$ and the observation at $\x_i=[x_{i,1},\ldots,x_{i,M}]^\top$ given by  $y(\x_i)=y_i=\sum_{j=1}^{M}w_{i,j}m_{i,j}$, this covariance is
\begin{align}
\label{eq:Kfc}
  K_{fy}(x, \x_i) &= \expectation{(f(x)-\mu_f(x))(y_i-\mu_{y}(\x_i)) } \\
  &= \sum_{j=1}^{M} w_{i,j} K_f(x,x_{i,j})\qquad \nonumber \\
  &={ \w^\top K_f(x,\x_i) \qquad}\nonumber
\end{align}
Denoting $\y=[y(\x_1),\ldots,y(\x_N)]^\top$, the predictive posterior is:
\begin{align}
  \label{eqn:responsePredictiveMean}
  \expectation{f(x)|\y} &= K_{fy} K_y^{-1}\y\\
  \label{eqn:responsePredictiveCov}
  \mathrm{Var}(f(x)|\y) &= K_f(x,x)-  K_{fy} K_y^{-1}K_{yf}
\end{align}
thus, the reconstruction of the latent process $f$ can be computed in closed-form (we used the notation $K_{fy}^\top=K_{yf}$).

\subsection{The Moore-Penrose solution is a particular case of GPMM}

Without loss of generality, let us consider that observations $y_i$ were taken at a grid $\x_\text{grid}=[x_1,\ldots,x_M]$ were only a few weights are nonzero per observation. Furthermore, without evidence for spatial correlation of the latent process $f(\cdot)$, its covariance matrix is the identity multiplied by $\sigma_f^2$ (signal power)---i.e., $\mathbf{K}(\x_\text{grid},\x_\text{grid})=\sigma_f^2\mathbf{I}_M$. Consequently, denoting the mixing weights by $\W=[\w_1,\ldots,\w_N]$ and $\x$ the input of the observed process, from eqs. \eqref{eq:Ky}-\eqref{eq:Kfc} the covariances are given by
\begin{align}
  K_{y}(\x,\x) &= (\sigma_\epsilon^2+\sigma_f^2)\W^\top \W +\sigma^2_\eta\mathbf{I}_M\\
  K_{fy}(x, \a) &= \sigma_f^2 \W^\top.
\end{align}
Finally, introducing the above two expressions in the posterior mean in eq.~\eqref{eqn:responsePredictiveMean} gives the solution to the linear system
\begin{align}
  \hat{f}(\x) &=  \frac{\sigma_f^2}{\sigma_\epsilon^2+\sigma_f^2}  \W^\top \left(   \W^\top  \W +\frac{\sigma^2_\eta}{\sigma_\epsilon^2+\sigma_f^2}\mathbf{I}_M\right)^{-1} \y.
\label{eq:GP_linearsol}
\end{align}

The connection between solutions to linear systems and GP models is therefore established: when the noise variances are negligible w.r.t. the signal power $\left(\sigma_f^2\gg\sigma_\epsilon^2,\sigma_\eta^2\right)$ the ratios  in eq.~\eqref{eq:GP_linearsol} ${\sigma_f^2}/{(\sigma_\epsilon^2+\sigma_f^2)}\rightarrow 1$ and ${\sigma_\eta^2}/{(\sigma_\epsilon^2+\sigma_f^2)}\rightarrow 0$, and the Moore-Penrose inverse is obtained. On the contrary, when the noise variance $\sigma_\epsilon^2$ is large the estimate decays to zero, or \textit{reverts to the prior}, since the measurements are not reliable. Furthermore, when the observation noise $\sigma^2_\eta=0$, eq.~\eqref{eq:GP_linearsol} is equivalent to the ordinary least squares and when $\sigma^2_\eta>0$ to the regularised least squares (ridge regression).

 Finally, we emphasise that, unlike the Moore-Penrose method, the proposed GPMM jointly infers the mixing weights and the complete latent process, while also providing a measure of uncertainty given by the variance in eq. \eqref{eqn:responsePredictiveCov}.

\section{Simulations}

The proposed GPMM model was validated using three datasets: A heart-rate time series, a smooth function generated by a GP with square exponential (SE) covariance kernel, and the Heaviside step function. All three experiments consisted in recovering the latent process from noisy mixtures of measurements by fitting GPMM to the observations and then computing the predictive posterior as described in Section \ref{sec_posterior}, where the learnt weights were constrained to have unit $L_1$-norm, positive entries and be symmetric to avoid redundant solutions. The proposed GPMM was compared to the standard GP that considers the observations as a single measurement in a single location, which is the common practice in sensing applications.

\subsection{Smooth synthetic signal}
\begin{figure}[t]
  \centering
\includegraphics[width=.49\textwidth]{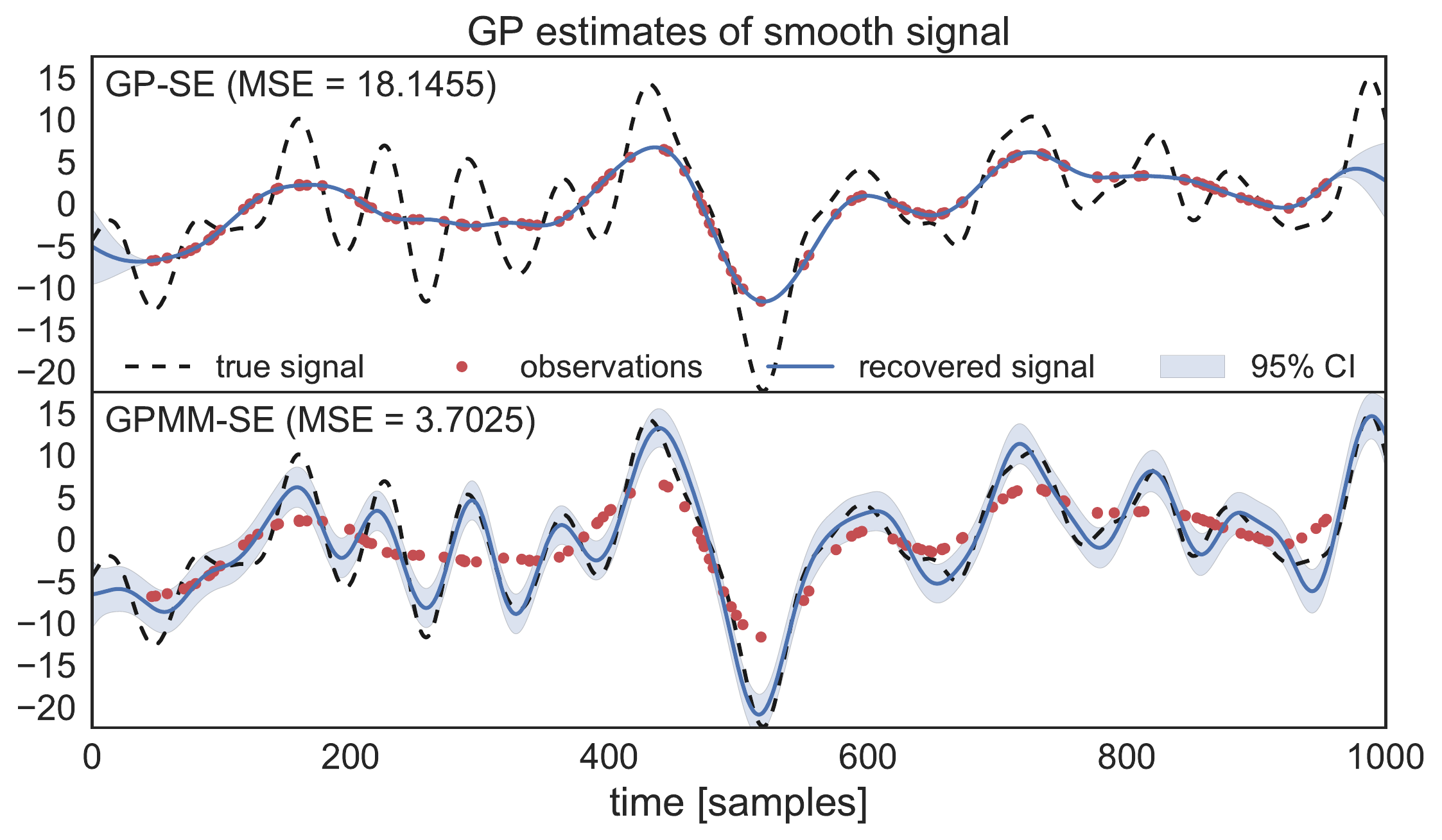}
\caption{Recovering a synthetic signal (dashed) from mixture of measurements (red): standard GP-SE (top, blue) and the proposed GPMM-SE (bottom, blue).}
\label{fig:smooth}
\end{figure}

This first toy example considered a draw from a GP with SE covariance as the latent function and 120 observations with 7 measurements per observation. Fig. \ref{fig:smooth} shows the GP estimates and their mean square error (MSE) for both the standard GP with SE kernel (termed GP-SE) and the proposed GPMM with SE kernel (termed GPMM-SE). Notice how GP-SE fails to recover all the extrema of the latent process and adjust very tightly to the observations, this is because the convolution performed by the sensor smooths out the extrema in the observations which are then trusted as true values by GP-SE. Conversely, the GPMM-SE was able to recover all the extrema, place appropriate error bars on the latent function and report a lower estimation error.

\subsection{Heart-rate signal}

\begin{figure}[t]
  \centering
  \includegraphics[width=.49\textwidth]{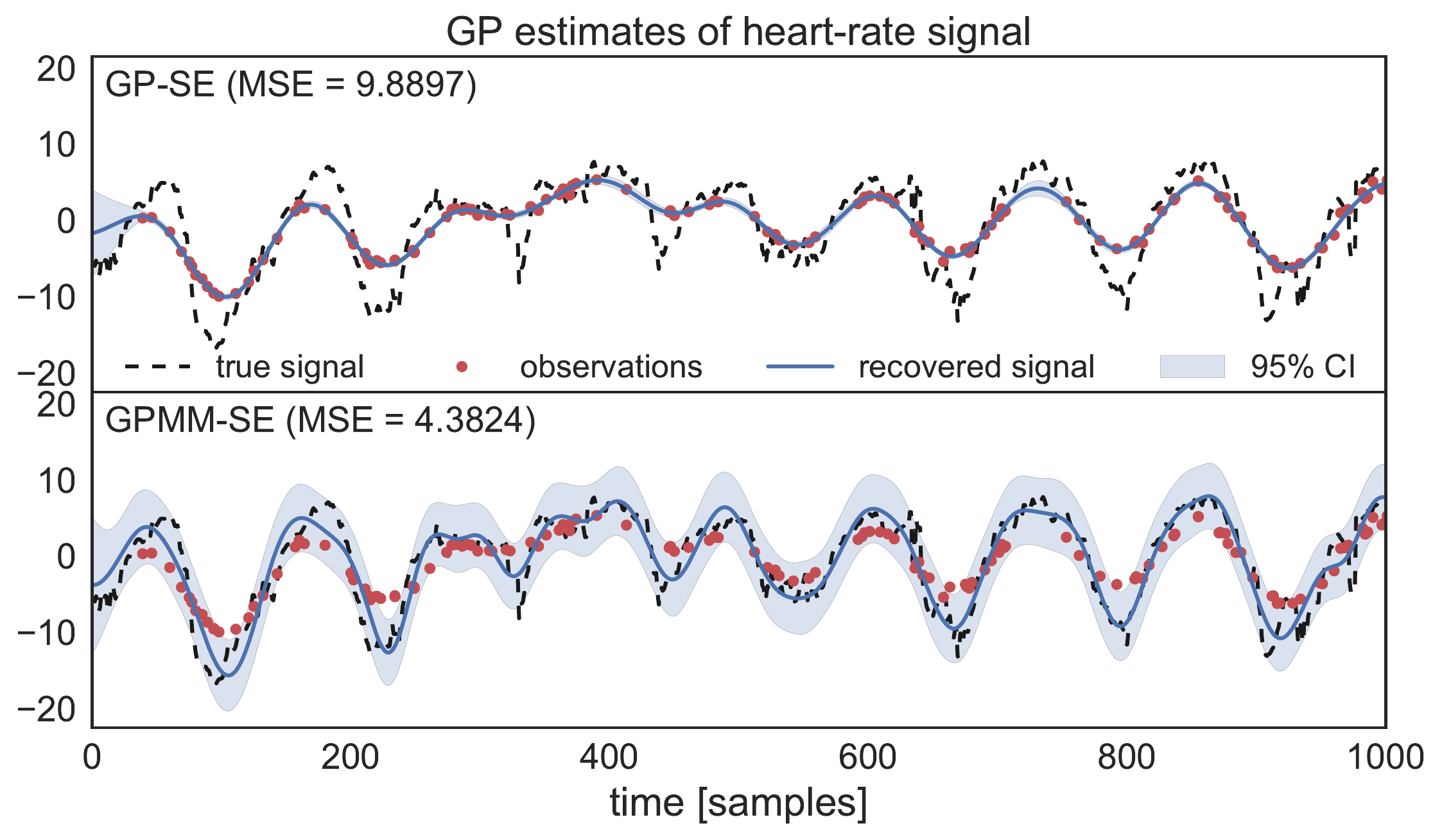}
  \caption{Recovering a heart-rate signal (dashed) from mixture of measurements (red): standard GP-SE (top, blue) and the proposed GPMM-SE (bottom, blue).}
  \label{fig:hr}
\end{figure}

Instantaneous-frequency estimation is performed averaging over a time window thus motivating the use of the proposed GPMM. We used a heart-rate time-series from the MIT-BIH Database \cite{Goldberger} (\href{http://ecg.mit.edu/time-series/}{\texttt{ecg.mit.edu/time-series}}) and constructed a lower-resolution version of it composed by a mixture of measurements. Fig. \ref{fig:hr} shows the recovery of a heart-rate signal from such observations for both the GP-SE and the proposed GPMM-SE, where the GPMM-SE again outperforms the GP-SE in terms of estimation MSE and uncertainty representation. We used 240 observations with 7 measurements each.

\subsection{Heaviside step function}
\begin{figure}[t]
  \centering
\includegraphics[width=.49\textwidth]{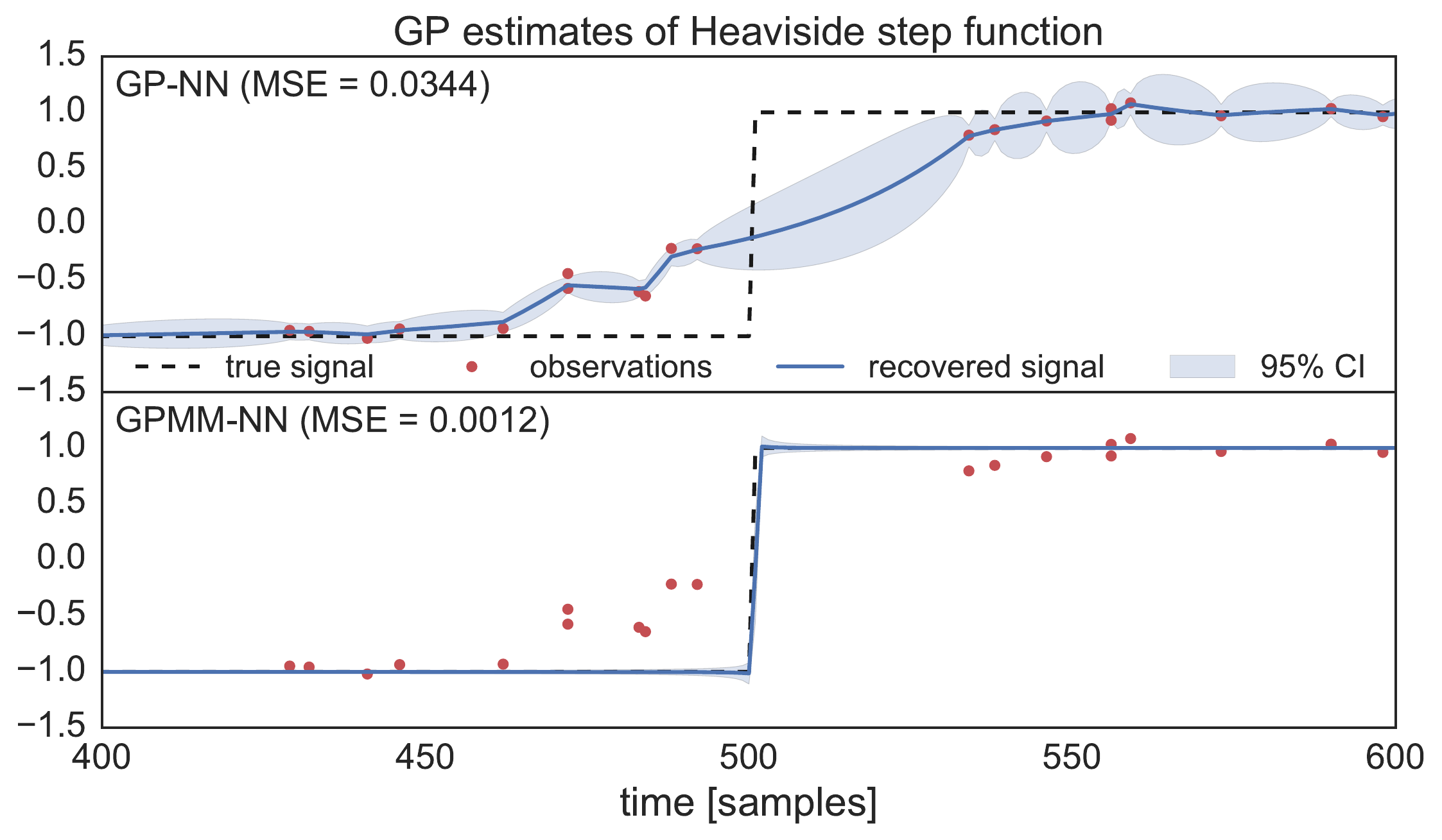}
  \caption{Recovering a step function (dashed) from mixture of measurements (red): standard GP-SE (top, blue) and the proposed GPMM-SE (bottom, blue).}
  \label{fig:heaviside}

\end{figure}

Motivated by edge-detection applications, we then considered the Heaviside step function. To cater for discontinuous signals, we used the neural network (NN) kernel \cite{will_inf} and implemented both the standard GP-NN and the proposed GPMM-NN to recover the latent step function. Observe in \ref{fig:heaviside},  how GPMM-NN successfully recovered the discontinuity with low variance. We used 120 observations with 7 measurements each.

\subsection{Comparing recovered signals in spectral terms}
\begin{figure}[t]
  \centering
\includegraphics[width=.49\textwidth]{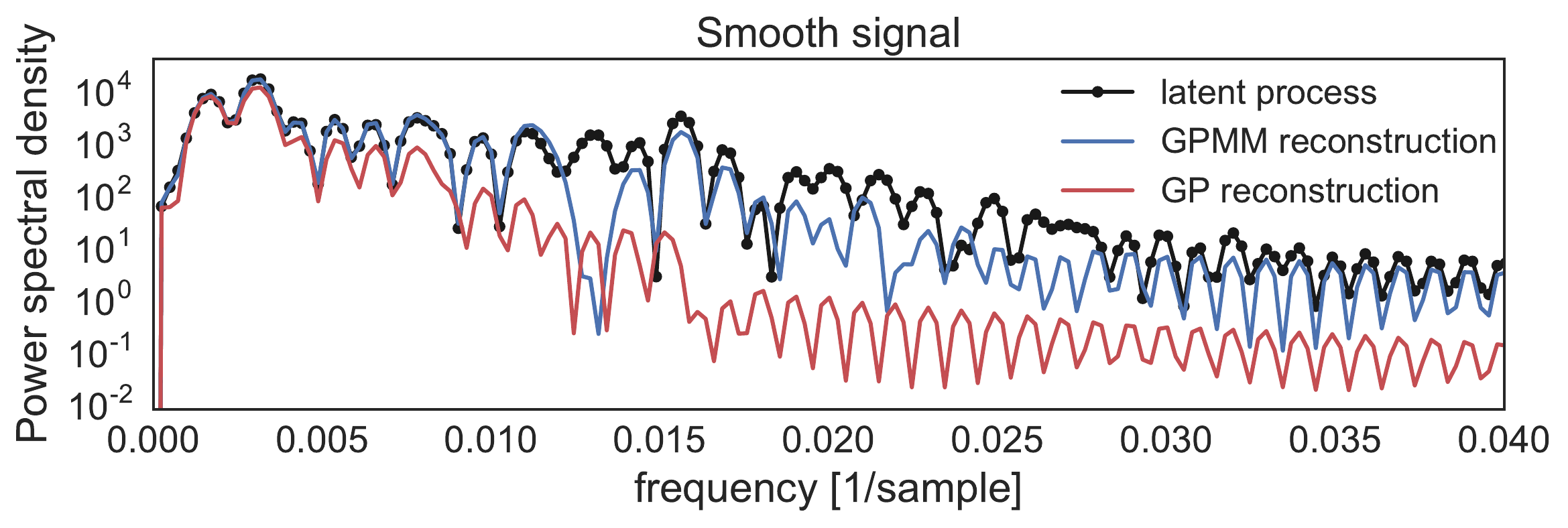}\\
\includegraphics[width=.49\textwidth]{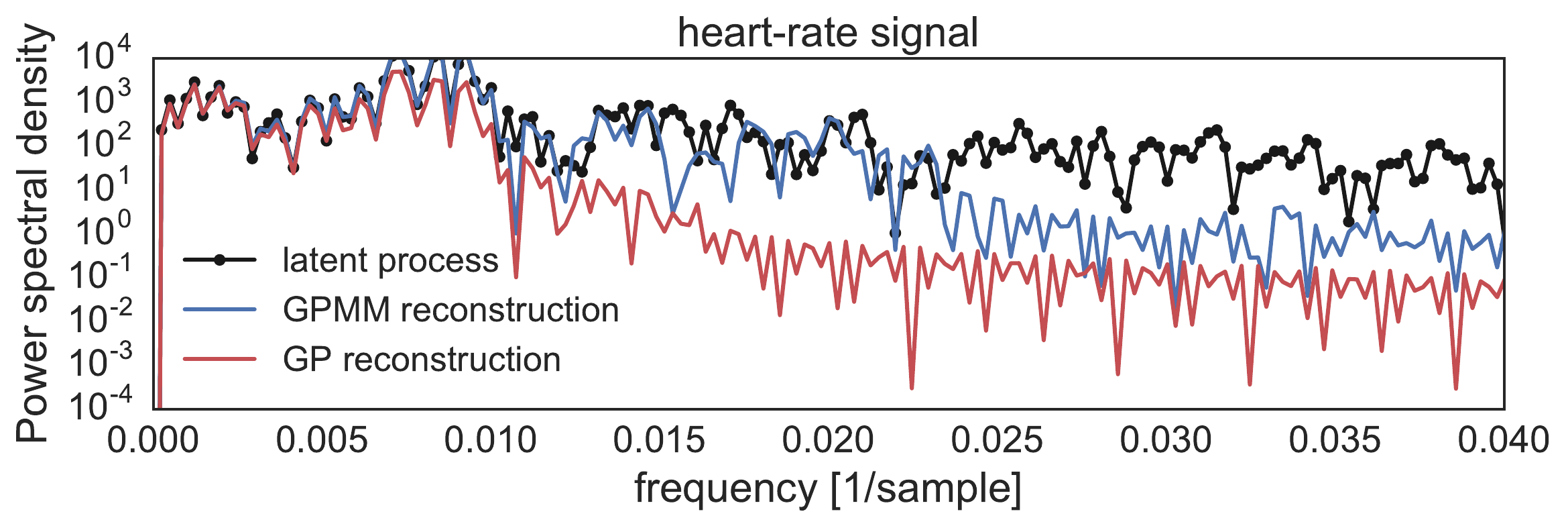}\\
\includegraphics[width=.49\textwidth]{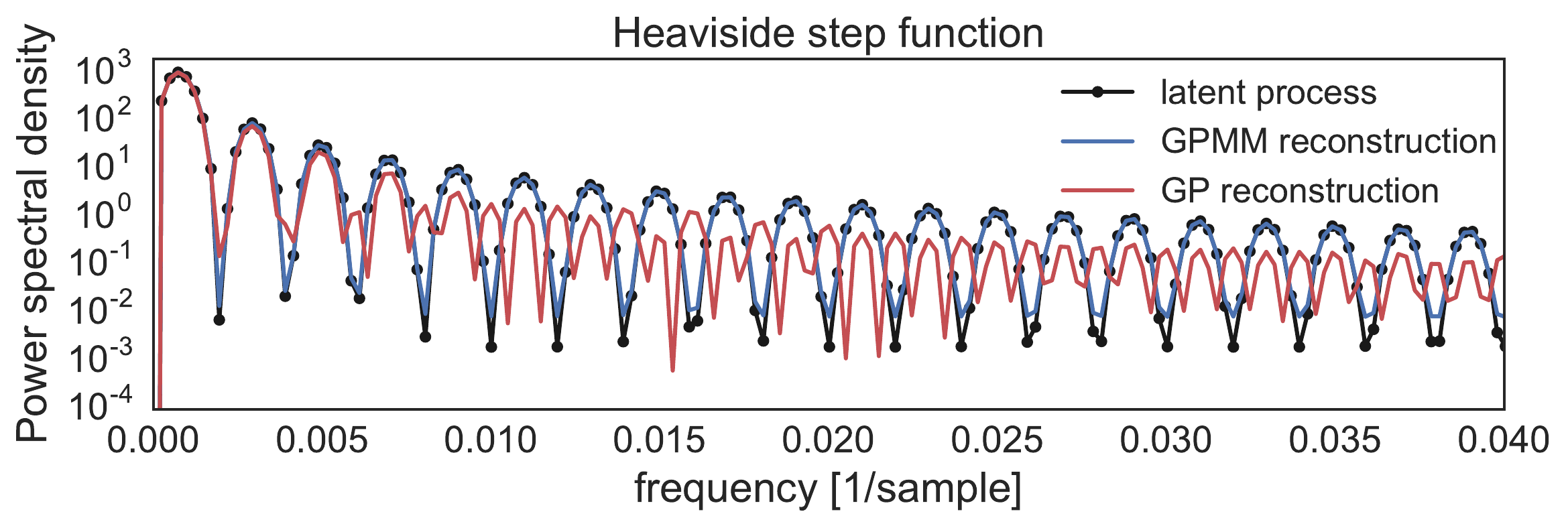}
  \caption{Power spectral density of true latent signal (black), proposed GPMM (blue) and standard GP (red) for all datasets considered: smooth signal (top), heart-rate signal (middle) and step function (bottom).}
  \label{fig:spectra}

\end{figure}

 Our aim is to recover spatial structure in the latent process, in this sense, Fig. \ref{fig:spectra} shows the PSDs for the smooth, heart-rate, and step function from top to bottom, observe how the GPMM (blue) was able to better recover the spectrum in all the experiments considered unlike the standard GP (red), where the latter failed to identify the spectral content of the latent process that was removed by the convolution sensor.

\section{Conclusions}
We have proposed the Gaussian process mixture of measurements (GPMM) to address the problem of recovering a latent signal from a noisy mixture of measurements of such signal, a common setting in sensing applications. Our contributions are (i) to model the latent process and the mixture-of-measurements process as jointly Gaussian, (ii) fitting GPMM, including the mixing weights, and deriving the posterior distribution of the latent function in closed form, (iii) interpreting the solution of the underdetermined linear system generated by the sensing application as a particular case of GPMM, and (iv) validating GPMM against the standard GP for synthetic and real-world signals, where the reconstruction accuracy of GPMM was evidenced both in the time and frequency domains.

\bibliographystyle{IEEEbib}
\bibliography{libraryGPMM}

\end{document}